\documentclass{article}
\usepackage{microtype}
\usepackage{graphicx}
\usepackage{subfigure}
\usepackage{booktabs} 
\usepackage{array}
\usepackage{enumitem}
\usepackage{hyperref}

\usepackage[accepted]{icml2023}

\usepackage{amsmath}
\usepackage{amssymb}
\usepackage{mathtools}
\usepackage{amsthm}
\usepackage{algorithm}
\usepackage{listings}
\usepackage[capitalize,noabbrev]{cleveref}
\theoremstyle{plain}

\theoremstyle{definition}

\theoremstyle{remark}

\newcommand{\method}{\texttt{DiversiGATE}~}
\newcommand{\methodns}{\texttt{DiversiGATE}} 
\newcommand{\selfmethod}{\texttt{SelfLearner}~}
\newcommand{\selfmethodns}{\texttt{SelfLearner}} 
\newcommand{\dataset}{{\texttt{GSM8K}~}}

\definecolor{question_color}{rgb}{0.3, 0.4, 0.65}
\definecolor{answer_color}{rgb}{0.6, 0.3, 0.45}

\definecolor{example_color}{rgb}{0, 0.4, 1}
\definecolor{green_color}{rgb}{0.3, 0.6, 0.2}
\definecolor{prompt_color}{rgb}{0.3, 0.3, 0.3}
\definecolor{notation_color}{rgb}{0.8, 0.2, 0.6}
\definecolor{baseline}{rgb}{0.1, 0.4, 0.5}
\definecolor{gray}{rgb}{0.3, 0.3, 0.3}
\definecolor{darkgreen}{rgb}{0.2, 0.6, 0.2}
\definecolor{darkred}{rgb}{1, 0, 0}

\newcommand{\rom}[1]{\uppercase\expandafter{\romannumeral #1\relax}}

\usepackage[textsize=tiny]{todonotes}

\icmltitlerunning{\method Framework for Reliable LLMs}

\begin{document}

\twocolumn[
\icmltitle{\methodns: A Comprehensive Framework for Reliable\\ Large Language Models}





\begin{icmlauthorlist}
\icmlauthor{~~Shima Imani}{a}
\icmlauthor{~~~~~~Ali Beyram}{b}
\icmlauthor{~~~~~~Harsh Shrivastava}{a}
\vspace{0.5mm}\\
{Microsoft Research, Redmond, USA$^1$} ~~~{ModelFarm.ai$^2$}\\\vspace{1mm}
\textit{Contact: shimaimani@microsoft.com}
\end{icmlauthorlist}




\vskip 0.3in
]



\begin{abstract}

In this paper, we introduce \methodns, a unified framework that consolidates diverse methodologies for LLM verification. The proposed framework comprises two main components: Diversification and Aggregation which provide a holistic perspective on existing verification approaches, such as Self-Consistency, Math Prompter and WebGPT. Furthermore, we propose a novel `\selfmethodns' model that conforms to the \method framework which can learn from its own outputs and refine its performance over time, leading to improved accuracy. To evaluate the effectiveness of \selfmethodns, we conducted a rigorous series of experiments, including tests on synthetic data as well as on popular arithmetic reasoning benchmarks such as \dataset. Our results demonstrate that our approach outperforms traditional LLMs, achieving a considerable 54.8\%$\rightarrow$61.8\% improvement on the \dataset benchmark.

\end{abstract}

\section{Introduction}
In recent years, Large Language Models (LLMs) have achieved significant advancements in various Natural Language Processing (NLP) tasks~\cite{devlin2018bert,brown2020language,raffel2020exploring,gao2020making,cobbe2021training,reynolds2021prompt,liu2021makes,geva2021did,patel2021nlp,rae2021scaling,chowdhery2022palm,thoppilan2022lamda,wang2022self,wei2022chain,srivastava2022beyond,kojima2022large,liu2023pre,imani2023mathprompter}. However, despite these accomplishments, LLMs continue to face challenges in generating accurate and reliable outputs, often producing false or incorrect information, referred to as hallucinations. This issue has spurred researchers to investigate novel techniques and methodologies to enhance the reliability and accuracy of LLM-generated outputs.

To address the concerns on reliability of LLMs, a popular approach by~\citet{wang2022self}, which seeks to generate `high-confidence' answers through a technique known as `self-consistency'. Their method involves running the LLM multiple times on a given input to generate a diverse set of candidate outputs that encompass various reasoning paths. The answers are then aggregated by selecting the output that demonstrates the most consistent answer, thus improving the reliability and accuracy of the model's output.

On similar lines, another interesting approach called `MathPrompter' by~\citet{imani2023mathprompter} showed improved results on mathematical reasoning tasks. This paper introduces the concept of creating an algebraic template of the given question and then querying the LLM to generate diverse solutions to the template query, which can even be a Pythonic function or an algebraic expression. The authors then use random input values as arguments and evaluate the LLMs until a consensus is reached. Then a majority voting mechanism is utilized to answer the final query. This approach reinforces the reliability of the model's output by considering the consensus among diverse answering strategies.

Incorporating external references, such as Bing or Google search, can also enhance the accuracy and reliability of LLM outputs, as demonstrated in the `WebGPT' model by~\citet{nakano2021webgpt}. By utilizing these additional sources of information, LLMs gain access to contextual references that increase confidence in their outputs. 


The methods discussed above were independently developed and share a common goal of providing high-confidence results to the users and thus increase the reliability of LLMs. In this paper, we present \methodns, which consolidates a diverse array of such existing methodologies into a single unified framework. Our framework comprises two main components: Diversification and Aggregation, collectively referred to as `\method' (\textbf{Diversi}fication and aggre\textbf{GAT}ion for v\textbf{E}rification). Our analysis reveals that  approaches, such as self-consistency, MathPrompter, WebGPT and others, share similarities and primarily differ in their implementation of these two components.

Based on the \method framework's two-step process of diversification followed by aggregation, we developed a novel approach, called \selfmethod that can improve its accuracy on a dataset over time. We experimentally demonstrate the effectiveness of our approach on the \dataset dataset. Our research aims at enhancing the reliability and accuracy of LLM outputs for various natural language processing tasks. The contributions of this paper are twofold:

\begin{itemize}[leftmargin=*,nolistsep]
    \item Introduced \methodns, a general unified framework for increased reliability in LLM outputs.
    \item Developed \selfmethod model that is based on the \method framework, which can improve its accuracy on a dataset over time. 
\end{itemize}


\section{Method}
\begin{figure}[h]
\centering
\includegraphics[width=80mm]{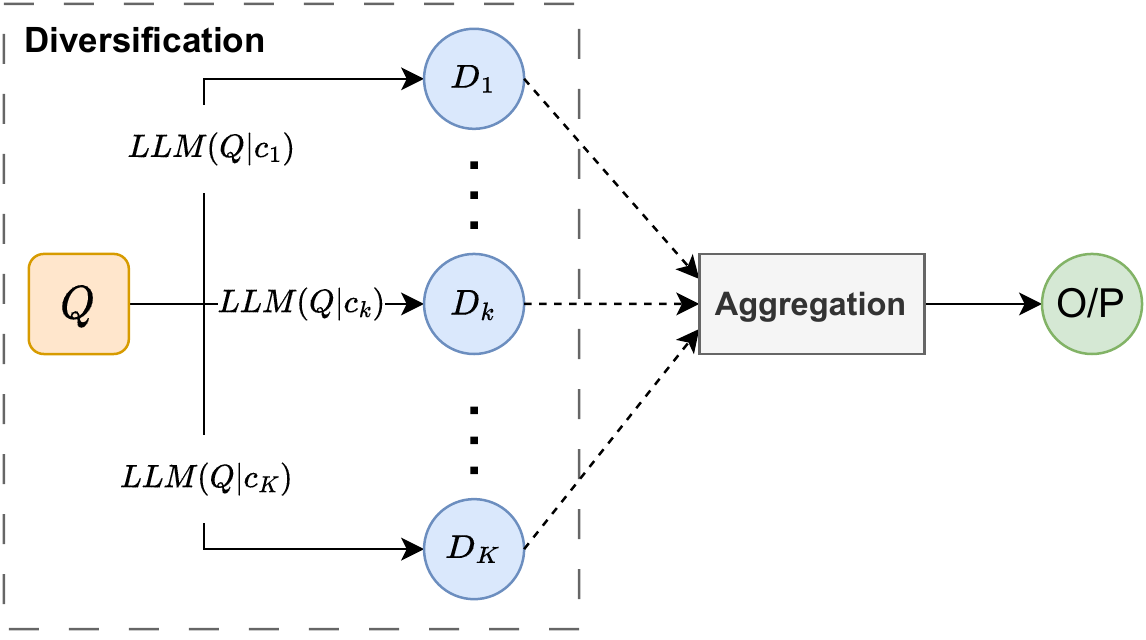}
\caption{\small \textit{\method framework}. Proposed unified  framework consisting of two steps, namely, `Diversification' module followed by an `Aggregation' module. The given input query $\mathcal{Q}$ is conditioned on diverse contexts $\{c_1,\cdots,c_K\}$ when passing through LLMs. These contexts can be different LLM settings such as temperature or different input prompts like in zero-shot or few-shot settings. This generates a diverse set of 
outputs $\{D_1,\cdots,D_K\}$ which are then aggregated to give the final output answer. A single phase of the framework is depicted above. A multi-phase (or chaining) of the above framework can represent many existing approaches designed to increase LLM's reliability.}
\label{fig:diversigate_flow}
\end{figure}

\begin{figure*}[t!]
\centering
\includegraphics[width=175mm]{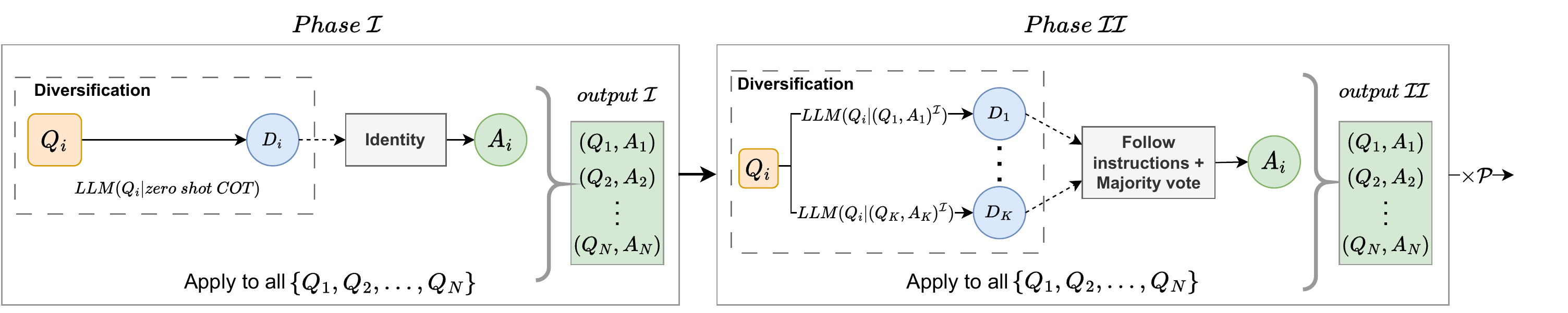}
\vspace{-4mm}
\caption{\small \textit{\selfmethod model}. It uses a chain of phases (or instances) of the \method framework, with each phase utilizing the output of the previous one with their own customized `Diversification' and `Aggregation' modules.}
\label{fig:self_learner_flow}
\end{figure*}

\method framework has two modules: The `Diversification' module aims to generate a diverse set of outputs for the input query by using different contexts. The `Aggregation' module combines these outputs to produce a single, more accurate result. Fig.~\ref{fig:diversigate_flow} shows a high-level overview of the framework.

\textit{Diversification module}. To generate a diverse set of outputs from the LLM, several techniques can be employed, such as few-shot learning, context manipulation, and adjusting the LLM settings (e.g. temperature setting). Few-shot learning involves providing different exemplars to the LLM, which can lead to the generation of diverse responses. For instance, in the `self-consistency' technique by~\citet{wang2022self}, the authors demonstrate the use of few-shot learning to generate a diverse set of outputs. Another way of generating diverse outputs can be to provide different contexts to the input query. For instance, `MathPrompter' by~\citet{imani2023mathprompter}, utilizes contexts such as `answer based on Python code' or `answer by an algebraic formula' to create diverse responses for mathematical reasoning tasks. `WebGPT' by~\citet{nakano2021webgpt} leverages search engines to generate diverse results for the input queries. It is important to emphasize that in each of these cases, distinct approaches are employed to generate a diverse set of outputs, which serves as an important element in achieving a higher confidence level.

\textit{Aggregation module}. This module decides a final consensus output from the diverse set of results obtained from the `Diversification' module. Some common aggregation methods include filtering process, majority voting and summarization. For example, the `self-consistency' method employs filtering and majority voting to obtain the final result, while `MathPrompter' aggregates the results by using majority voting criterion. `WebGPT', on the other hand, uses summarization as its aggregation method.

It is an important observation that in prior approaches, \textbf{no learning} occurs in the `Diversification' and `Aggregation' modules. In the next section, we introduce a novel method, \selfmethodns, which has the potential to learn and thereby improve the outcomes of each of the modules.

\subsection{\selfmethod model}
There are two complementary paradigms that are popularly used to query LLMs: zero-shot learning and few-shot learning. We briefly review them before explaining our model.

\textit{Zero-shot learning}. This is a machine learning technique that enables the recognition of objects or the performance of tasks without having seen any examples of the target class during training. It is particularly useful in NLP domain for querying LLMs to improve output results. 
One such technique, termed as \textit{Zero-shot Chain of Thought (CoT)}, proposed by~\citet{wei2022chain}, proposed appending the phrase `Let's think step-by-step' as a prompt to the  input query of LLM. They found that it helps in generating a sequence of concise statements that emulate the cognitive reasoning process an individual might utilize when addressing a particular task. Notably, this approach has demonstrated a substantial enhancement in model performance across an array of multi-step reasoning tasks.

\textit{Few-shot learning}. This machine learning  technique allows a model to learn and generalize from a small number of examples. It is helpful when there are limited labeled data for a specific task or class. Unlike zero-shot learning, few-shot learning provides the model with a few examples to learn the target class.~\citet{brown2020language} extended this technique to query LLMs.

Both these learning paradigms have demonstrated promising results across diverse applications. The \selfmethod model combines their strengths to get better performance. In this work, we primarily focused on reasoning tasks, but our approach can be generalized to a wide range of tasks. Proposed \selfmethod conforms to the \method framework. We experimented with the two phases design shown in Fig.~\ref{fig:self_learner_flow}, but we can potentially design more phases as needed. We use an example to illustrate the internal module designs. Consider the following sample question `{\color{gray} $Q_i$}' from the \dataset dataset.
\begin{quotation}
     \noindent{\color{gray} $\mathcal{Q}_i$: \textcolor{question_color}{Betty is saving money for a new wallet which costs \$100. Betty has only half of the money she needs. Her parents decided to give her \$15 for that purpose, and her grandparents twice as much as her parents. How much more money does Betty need to buy the wallet?}}
\end{quotation} 

\textit{Phase $\mathcal{I}$-Diversification module}. It is a zero-shot CoT method to generate reasoning outputs for each question in the dataset. 
By appending `Let's think step-by-step' to the input prompt, the following output `{\color{gray} $D_i$}' is generated.
\begin{quotation}
     \noindent{\color{gray} $D_i$: \textcolor{answer_color}{Betty has \$15 from her parents. 2. Betty's grandparents gave her twice as much as her parents, so that's \$30. 3. Betty has a total of \$45. 4. Betty needs \$100 to buy the wallet. 5. Betty needs \$55 more to buy the wallet.}}
\end{quotation}

\textit{Phase $\mathcal{I}$-Aggregation module}. We use an identity block $A_i=D_i$, thus the output of Phase $\mathcal{I}$ are the pairs $(Q_i, A_i)$.

\textit{Phase $\mathcal{II}$-Diversification module}. The output pairs from the \textit{Phase $\mathcal{I}$} are used as a one-shot prompts to generate multiple answers for each question. 
For our sample query $Q_i$, we use $K=20$ one-shot prompts that act as diverse contexts. These prompts are the output pairs $(\mathcal{Q}, \mathcal{A})^\mathcal{I}$ generated from \textit{Phase $\mathcal{I}$}. We show a sample context $k$ below

\begin{quotation}
     \noindent{\color{gray} $\mathcal{Q}_k^\mathcal{I}$: \textcolor{gray}{James writes a 3-page letter to 2 different friends twice a week. How many pages does he write a year?}}

     \noindent{\color{gray} $\mathcal{A}_k^\mathcal{I}$:
     \textcolor{gray}{1. James writes a 3-page letter to 2 different friends twice a week. 2. There are 52 weeks in a year. 3. Therefore, James writes a total of 312 pages a year (3 pages x 2 friends x 52 weeks).}}
\end{quotation}

Incorporating this one-shot prompt as a context for $\mathcal{Q}_i$ yields the following response $LLM(\mathcal{Q}_i|(\mathcal{Q}_k,\mathcal{A}_k)^I)$ as.

\begin{quotation}
     \noindent{\color{gray} $\mathcal{D}_k$: \textcolor{answer_color}{Betty needs \$55 more to buy the wallet. She already has \$50, and her parents gave her \$15 and her grandparents gave her twice as much, which is \$30. Therefore, she needs \$55 more to buy the wallet.}}
\end{quotation}

This process is performed for all the questions $\{\mathcal{Q}_1,\cdots,\mathcal{Q}_N\}$ in the dataset, with each $\mathcal{Q}_i$ having $\{\mathcal{D}_1,\cdots,\mathcal{D}_K\}^i$ diverse answers.

\textit{Phase $\mathcal{II}$-Aggregation module}. In order to obtain a consensus among the $K$ diverse answers to finalize output answer, the `follow instructions' along with a `majority vote' strategy is used. The context in the Diversification module uses the pair $(\mathcal{Q},\mathcal{A})^\mathcal{I}$, where $\mathcal{A}^\mathcal{I}$'s are generated using the zero-shot CoT which leverages a step-by-step context for generating results (from \textit{Phase $\mathcal{I}$}). The `follow instruction' filter chooses answers based on the presence of the word `Step'. For the example $\mathcal{Q}_i$, we only consider the $\mathcal{D}_k$'s where the $\mathcal{A}_k^\mathcal{I}$ has the word `Step' present. The filtering process is followed by counting the majority vote of the remaining $\mathcal{D}_k$'s to obtain the final answer.

\begin{quotation}
     \noindent{\color{gray}  $\mathcal{A}_i$: \textcolor{answer_color}{Step 1: Calculate how much money Betty has. Betty has half of the money she needs for the wallet, which is \$50. Her parents gave her \$15 and her grandparents gave her twice as much, which is \$30. That means Betty has a total of \$95. Step 2: Calculate how much more money Betty needs to buy the wallet. Betty needs \$100 to buy the wallet. She already has \$95, so she needs \$5 more. Therefore, Betty needs to save \$5 more to buy the wallet.}}
\end{quotation}
The output of this \textit{Phase} is an updated set of pairs $(\mathcal{Q},\mathcal{A})^\mathcal{II}$. We anticipate that the model's performance will enhance with each subsequent \textit{Phase}.



\selfmethod enables the generation of a diverse set of answers for each question without requiring any manual creation of prompts. We emphasize that our model operates in an unsupervised manner, i.e. we do not discard any output pairs, regardless of their correctness.

\section{Experiments}


\begin{table*}[t]
\centering
\small
\begin{tabular}{|c|c|c|c|c|}
\hline
\rowcolor[HTML]{DAE8FC} 
\textbf{Method} & \textbf{Diversification context} & \textbf{Aggregation module} & \textbf{Multiplication} & \textbf{Division} \\ \hline
Zero-shot         & Identity                 & Identity      & 85.80\% & 78.40\%  \\ \hline
One-shot          & {[}random `correct' prompt{]}×1                 & Identity      & 88.95\% & 93.70\% \\ \hline
\selfmethod     & {[}random prompt{]}×1  & Identity ($\mathcal{I;II}$)     & 88.70\% & 85.70\% \\ \hline
One-shot Ensemble & {[}random  `correct' prompt{]}×20 & Maj-vote & 90.00\% & 96.20\% \\ \hline
\selfmethod     & {[}random prompt{]}×20 & Identity($\mathcal{I}$); Maj-vote($\mathcal{II}$) & 89.00\% & 92.40\% \\ \hline
\end{tabular}
\vspace{-2mm}
\caption{\small \textit{Synthetic data}. Accuracy results for the Multiplication and Division tasks are reported. We observe that the 2-\textit{Phase} \selfmethod outperforms the traditional zero-shot and one-shot learning approaches by a significant margin. We note that the results of \selfmethod are only second to the One-shot Ensemble learning which utilizes an ensemble of correct exemplars (or prompts) as opposed to the unsupervised prompts used by our method. }
\label{tab:synthetic_result}
\end{table*}

\subsection{Large Language Model}
We used the GPT-3 Davinci-text-003 engine for all our experiments. This engine is known for its advanced natural language processing capabilities and ability to generate human-like text, refer~\citet{brown2020language}.

\subsection{Datasets}

\subsubsection{Synthetic Dataset}\label{sec:synthetic_dataset}
We create datasets to evaluate LLM's capability to do simple multiplication and division tasks. Specifically, we utilized the following template for multiplication `{\textcolor{gray}{$\mathcal{Q}_M$: What is the product of \texttt{A} and \texttt{B}?}}'. The values of \{\texttt{A,B}\}$\sim U(1, 100)$ were sampled uniformly at random. We then store the multiplication answer \texttt{M} = \texttt{A}$\cdot$\texttt{B} and use it to create a corresponding division task using the template  `{\textcolor{gray}{$\mathcal{Q}_D$: What is the result of \texttt{M} divided by \texttt{A}?}}'. We create $500$ questions for each task.


\subsubsection{\dataset Dataset}
This dataset comprises 8.5K high-quality linguistically diverse grade school math word problems that were created by human problem writers~\cite{cobbe2021training}. It is segmented into 7.5K training problems and 1K test problems. These problems require between 2 and 8 steps to solve and their solutions primarily involve performing a sequence of elementary calculations using basic arithmetic operations. 

\subsection{Results}
\subsubsection{Evaluation on Synthetic datasets}
Our objective is to examine the efficacy of \selfmethod approach when applied to elementary arithmetic operations like multiplication and division, refer Sec.\ref{sec:synthetic_dataset}. 
We have deliberately excluded addition and subtraction tasks from our analysis, as these operations are comparatively straightforward, and the accuracy of LLMs in handling such tasks is frequently close to one. For fair comparison, we employ a consistent seed to ensure that the random numbers generated for each of the methods listed below are identical.

\begin{itemize}[leftmargin=*,nolistsep]
    \item \textit{Zero-shot learning}. Baseline method that directly queries the LLM with questions $\mathcal{Q}$. Evaluates LLM$(\mathcal{Q})$.
    \item \textit{One-shot learning}. Performs one-shot learning with a random `correct' exemplar $\texttt{e}$ as a prompt. Evaluates LLM$(\mathcal{Q}|\texttt{e})$.
    \item \textit{One-shot Ensemble learning}. Utilizes $K$ random `correct' exemplars as one-shot prompts to answer a single query.  We used $K=20$ such prompts and used majority voting as the aggregation technique. Evaluates $\operatorname{maj-vote}\left(\text{LLM}(\mathcal{Q}|\texttt{e}_1),\cdots, \text{LLM}(\mathcal{Q}|\texttt{e}_K)\right)$. 
    \item \textit{\selfmethodns}. We use \textit{two-Phase} model as shown in Fig.~\ref{fig:self_learner_flow}. We use two different settings with varying number of $K=\{1, 20\}$ in the \textit{Phase $\mathcal{II}$}-Diversification module. 
\end{itemize}




Table~\ref{tab:synthetic_result} summarizes the accuracy results of different methods on the synthetic data. We see that the accuracy of one-shot learning is significantly higher than the zero-shot learning. This indicates that incorporating a single `correct' example prompt can greatly enhance the LLM's performance in solving arithmetic problems. \selfmethodns's $K=1$ setting (row 3) essentially emulates an \textbf{unsupervised} version of one-shot learning with the possibility that the input prompt might have an incorrect solution and hence we see a decrease in accuracy compared to the one-shot learning (row 2) method. Similar trend is observed for the $K=20$ setting and as expected, it works better than $K=1$ setting. \selfmethod achieves a significant improvement in comparison to zero-shot learning model of $3.2\%$ for the multiplication task and of $14.0\%$ for the division task. We want to highlight some interesting analysis here.
\begin{itemize}[leftmargin=*,nolistsep]
\item Even random one-shot prompts from the dataset can improve the accuracy. Diversification introduced by using one-shot prompts can improve the model's reliability. 
\item We observed that LLMs, when queried directly, are more prone to hallucination for the division task than multiplication which explains the accuracy difference in zero-shot results (row 1). For instance, when presented with the question {\textcolor{gray}{Q: What is $8265$ divided by $87$?}}, the LLM produces an erroneous output as $839080459769$. Although one-shot prompts reduce such hallucinations considerably.
\item We speculate that the higher accuracy of division compared to multiplication in one-shot settings are due to the fact that the outputs for division are, on average, smaller than those for multiplication tasks. Consequently, there is a reduced margin for error in generating smaller numbers.
\end{itemize}

\subsubsection{Ablation study on \dataset}

\begin{table}[]
\centering
\small
\begin{tabular}{|c|c|c|}
\hline
\rowcolor[HTML]{D6F5F3} 
{\color[HTML]{000000} \textbf{Phase}}                            & 1      & 2    \\ \hline
\cellcolor[HTML]{FFFFFF}{\color[HTML]{000000} \textbf{Accuracy}} & 54.8\% & 61.8\%  \\ \hline
\end{tabular}
\vspace{-2mm}
\caption{\small \textit{\selfmethod on \dataset (7.5K  $\mathcal{Q}$'s)}. We vary the total number of \textit{Phase} blocks and observe the change in performance. Limited to 2 blocks due to computational constraints.}
\label{tab:gsm8k_result}
\end{table}

\begin{table}[]
\centering
\small
\begin{tabular}{|c|c|c|c|c|c|}
\hline
\rowcolor[HTML]{FFE8E8} 
{\color[HTML]{000000} \textbf{Phase}}                            & 1& 2& 3& 4& 5\\ \hline
\cellcolor[HTML]{FFFFFF}{\color[HTML]{000000} \textbf{Accuracy}} & 56\% & 60\% & 63\% & 61\%& 64\%\\ \hline
\end{tabular}
\vspace{-2mm}
\caption{\small  \textit{\selfmethod on \dataset (100 $\mathcal{Q}$'s)}. We observe an increasing trend in accuracy with increasing the number of \textit{Phase} blocks, thus emphasizing the `self-learning' ability of our model. We want to point out that our model can saturate in terms of accuracy results w.r.t. the number of blocks. 
}
\label{tab:gsm8k_small_result}
\end{table}

In order to gauge the contribution of each \textit{Phase} of \selfmethodns, we evaluate our framework by incrementally adding these blocks. \textit{Phases $\mathcal{I,II}$} are designed as shown in Fig.~\ref{fig:self_learner_flow} and any additional phase blocks follow the design of \textit{Phase$\mathcal{II}$}. Tables~\ref{tab:gsm8k_result}\&\ref{tab:gsm8k_small_result} demonstrate improvement after each \textit{Phase}, showcasing the ability to refine its performance over time. Our approach essentially `\textit{learns by itself}' as it is completely unsupervised and it achieves an impressive increase in accuracy from $54.8\%\rightarrow 61.8\%$ on the challenging \dataset dataset. 


\section{Conclusion and Future Work}

We introduced \method which is a unified comprehensive framework that encompasses various techniques to increase reliability of LLMs such as Self-Consistency, MathPrompter, and WebGPT. The framework consists of two primary components, Diversification and Aggregation, which offers a systematic perspective on the developing approaches to increase LLM reliability for various tasks. Furthermore, we have developed a novel \selfmethod model within the \method framework which enables our model to improve based on its own outputs and successively enhance its performance over time in a completely unsupervised manner. To evaluate various aspects of our framework, we designed a new synthetic dataset along with demonstrating a significant accuracy jump 54.8\%$\rightarrow$61.8\% on the \dataset dataset.

Our on-going efforts are geared towards smarter designing of the `Diversification' and `Aggregation' modules. Our exploratory experimentation that uses LLMs as an Aggregation module shows promise. We are also actively evaluating on different LLM versions and model sizes.


\bibliography{example_paper}
\bibliographystyle{icml2023}



\appendix

\end{document}